\documentclass{article}

\usepackage{arxiv}

\usepackage[utf8]{inputenc} 
\usepackage[T1]{fontenc}    
\usepackage{hyperref}       
\usepackage{url}            
\usepackage{booktabs}       
\usepackage{amsfonts}       
\usepackage{nicefrac}       
\usepackage{microtype}      
\usepackage{lipsum}
\usepackage{graphicx}
\graphicspath{ {./images/} }

\title{Learning Grammar of Complex Activities via Deep Neural Networks}

\author{
 Mashaido Becky \\
  Khoury College of Computer Sciences \\
  Northeastern University \\
  Boston, MA 02115 \\
  \texttt{mashaido.r@northeastern.edu} \\
}

\begin{document}
\maketitle
\begin{abstract}
Motivated by the growing amount of publicly available video data on online streaming services and an increased interest in applications that analyze continuous video streams such as autonomous driving \cite{art1}, this technical report provides a theoretical insight into deep neural networks for video learning, under label constraints. I build upon previous work in video learning for computer vision, make observations on model performance and propose further mechanisms to help improve our observations. 
\end{abstract}


\section{Introduction and Motivation}
\subsection{What is the problem?}
Video learning is increasingly becoming a huge part of automating computer vision tasks however with the growing amount of publicly available video data, comes the problem of unstructured video data. I address this problem by building upon a deep neural network architecture (proposed by Zijia Lu\footnote{Zijia Lu: \url{https://www.khoury.northeastern.edu/people/zijia-lu/}}) that predicts the task of a video and actions happening in the video. Similar to multi-instance learning, video-level labels here are accessible, yet accurate action predictions require finding the frame-level labels. It is therefore relied upon this model to find the correspondence between the video-level and frame-level labels.

\subsection{Why is it interesting and important? Why is it hard? (E.g., why do naive approaches fail?)}
This task is applicable to many other topics such as weakly supervised action segmentation. Weakly supervised action segmentation in test time, must analyse the relation between a test video with all possible action sequences from all tasks. This is computationally complex and prone to error. The model in this project is therefore proposed to reduce the complexity by predicting the most relevant tasks and narrowing down the range of possible action sequences. To this end, we experiment with different network architecture, motion features and visual features, and different training strategies.

\subsection{Why hasn’t it been solved before? (Or, what’s wrong with previous proposed solutions? How does   mine differ?)}
In addition to points laid out in \textit{section 1.2}, past research has explored the use of a NeuralNetwork-Viterbi \cite{art1} algorithm as shown in \textit{Figure 1}, to generate frame labels. In this project however, our baseline model (Zijia Lu) depends on the attention mechanism, as shown in \textit{Figure 2}.

\begin{figure}[p]
\centering
\includegraphics[width=\textwidth]{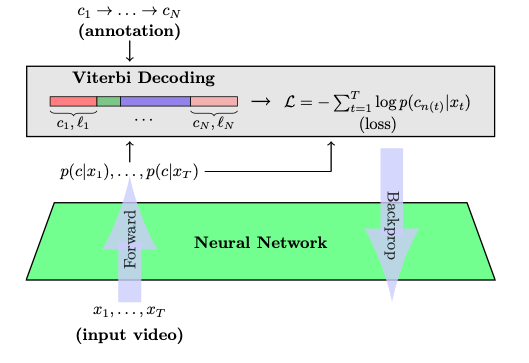}
\caption{Input video is forwarded through the network and the Viterbi decoding is run on the output probabilities. The frame labels generated by the Viterbi algorithm are then used to compute a framewise cross-entropy loss based on which the network gradient is computed.}

\includegraphics[width=\textwidth]{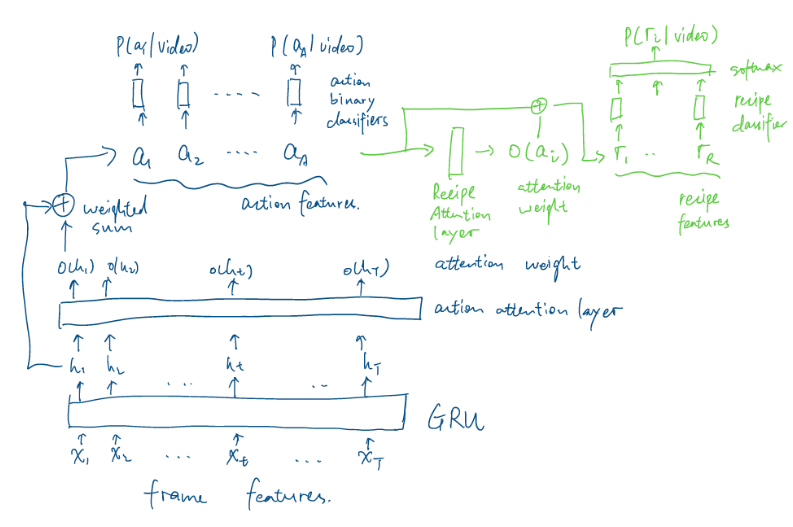}
\caption{Our architecture for the recipe action classifier, which uses an attention mechanism.}
\end{figure}

\subsection{What are the key components of my approach and results? Also include any specific limitations.}

\begin{enumerate}
\item Baseline model consists of a gated recurrent unit (GRU) and an attention (fully connected) layer as shown in \textit{Figure 2}
\item Video frames are fed into the GRU layer as features, and consequently through the action attention layer 
\item A weighted sum is generated and resulting action features later passed through the recipe attention layer
\item We also generate a temporal attention vector as an output of the action attention layer
\item We compute a softmax to turn the recipe logits (from the recipe attention layer) into probabilities
\item We use the temporal attention vetor from \textit{part 4} to analyze performance of our attention mechanism, as detailed in the \textit{results section 2.3}
\item We train and test our model on two dataset-features outlined in the  \textit{dataset section 2.1} and with each experiment, make observations on our accuracy
\item We try different methods to improve the accuracy of our model and note down our observations
\item Lastly, we make recommendations for next steps, based on our observations
\end{enumerate}

Our model performance did not seem to improve. We will get into this, by the end of this paper.

\section{Experiments and Results}
\subsection{Dataset}
This project involves working on the \href{https://serre-lab.clps.brown.edu/resource/breakfast-actions-dataset/}{breakfast} dataset which consist of diverse tasks:
\renewcommand\labelitemii{$\blacksquare$}
\begin{itemize}
\item    48 actions and 10 recipes related to breakfast preparation, performed by 52 different individuals in 18 different kitchens
\end{itemize}

We work with two sets of features:
\renewcommand{\labelitemii}{$\blacksquare$}
\begin{itemize}
\item    Motion: Improved Dense Trajectory + PCA; I3D Motion Feature
\item    Visual:  I3D Visual Feature; pretrained ResNet Feature
\end{itemize}

\subsection{Experimental Set-Up}
For all experiments, code was written in pytorch and numpy. A Neural network consisting of 1 hidden layer with 64 hidden neurons was used. Refer to \textit{Figure 3} for our network. We train using 10,000 epochs and a batch size of 16, to help with CUDA memory issues, though one could change these commands as seen in \textit{Figure 3}. Our network.py is where we set up our model and train.py is where we train our model. We then save the best checkpoint to test our model later.

\subsection{Results and Analysis}
During training we observed that our model overfits therefore we implemented various regularization techniques including dropout and weight decay, with results shown in \textit{Figures 4} and \textit{5}. We added these to the GRU layer and made sure to turn off dropout during testing. We also introduced our second set of features outlined in the \textit{dataset section 2.1} and performed experiments on both old (motion) and new (visual) set of features, as shown in \textit{Figures 4} and \textit{5}. We generated accuracy and loss plots to help visualize the performance of our model as shown in the sample \textit{Figures 7} and \textit{8}.

\includegraphics[scale=0.5]{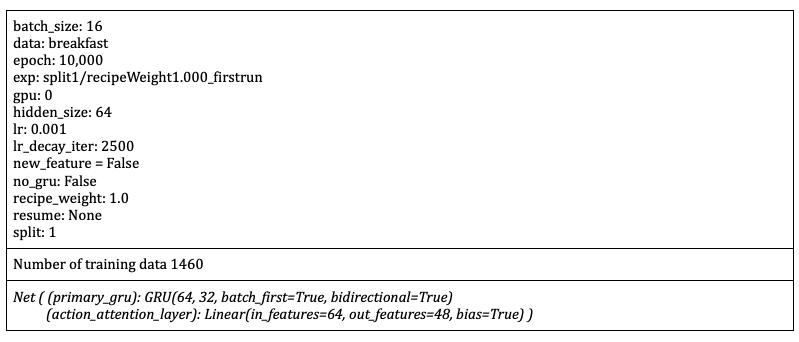}
\begin{figure}[h!]
\vspace*{-5mm}
\caption{Training commands and model set-up. Network consists of a gru and attention layer, with 64 hidden neurons.}
\end{figure}

\setlength{\arrayrulewidth}{1mm}
\setlength{\tabcolsep}{18pt}
\renewcommand{\arraystretch}{1.5}
\begin{tabular}{ |p{3cm}|p{3cm}|p{2cm}|p{2cm}|p{2cm}|  }
\hline
\multicolumn{5}{|c|}{Results on Motion Features} \\
\hline
Experiment Run & Experiment Setting & Test Recipe Accuracy & Test Action Accuracy & Test Action F1 \\
\hline
with dropout & rate p = 50\% & 0.466 & 0.909 & 0.27 \\
with dropout & rate p = 40\% & 0.471 & 0.906 & 0.269 \\
with dropout & rate p = 60\% & 0.487 & 0.914 & 0.295 \\
with weight decay i.e L2 regularization & value = 0.01 & 0.413 & 0.905 & 0.132 \\
\hline
\end{tabular}
\begin{figure}[h!]
\caption{Results: baseline and regularized model on motion (old set of) features.}
\end{figure}

\setlength{\arrayrulewidth}{1mm}
\setlength{\tabcolsep}{18pt}
\renewcommand{\arraystretch}{1.5}

\begin{tabular}{ |p{3cm}|p{3cm}|p{2cm}|p{2cm}|p{2cm}|  }
\hline
\multicolumn{5}{|c|}{Results on Visual Features} \\
\hline
Experiment Run & Experiment Setting & Test Recipe Accuracy & Test Action Accuracy & Test Action F1 \\
\hline
with dropout & rate p = 50\% & 0.551 & 0.918 & 0.299 \\
with dropout & rate p = 40\% & 0.521 & 0.916 & 0.29 \\
with dropout & rate p = 60\% & 0.497 & 0.917 & 0.28 \\
with weight decay & value = 0.01 & 0.469 & 0.922 & 0.202 \\
baseline without anything & N/A & 0.508 & 0.917 & 0.316 \\
with weight decay & value = 0.1 & 0.19 & 0.907 & 0 \\
with weight decay & value = 0.001 & 0.512 & 0.919 & 0.311 \\
with weight decay & value = 0.0001 & 0.504 & 0.919 & 0.314 \\
combining best dropout and best weight decay & P = 50\% and value = 0.0001 & 0.52 & 0.916 & 0.314 \\
\hline
\end{tabular}

\begin{figure}[h!]
\caption{Results: baseline and regularized model on visual (new set of) features.}
\end{figure}

After implementing regularization techniques we still encountered overfitting issues and therefore decided to take a look under the hood i.e fully understand our attention mechanism. Here, we analyzed our attention mechanism as documented on attention.py and visualized our attention distribution for each test video as shown in the sample \textit{Figure 9}. We also compute a quantitative score to help measure the quality of our attention as outlined in attention.py, using \textit{Figure 6} formula below.

\begin{equation}
Score_{v} = \frac{1}{A} \sum_{i=1}^{A}\frac{1}{T}\cdot l_{i}\cdot P(a_{i})
\end{equation}
\begin{figure}[h!]
\caption{Computing a quantitative score to measure the quality of attention for each test video, using the one-hot form for the ground truth label in each frame and the corresponding action attention vector as mentioned in section 1.5. $P(a_{i})$ is the attention vector for action $a_{i}$ and $l_{i}$ is the one-hot label for action $a_{i}$.}
\end{figure}

\includegraphics[scale=0.48]{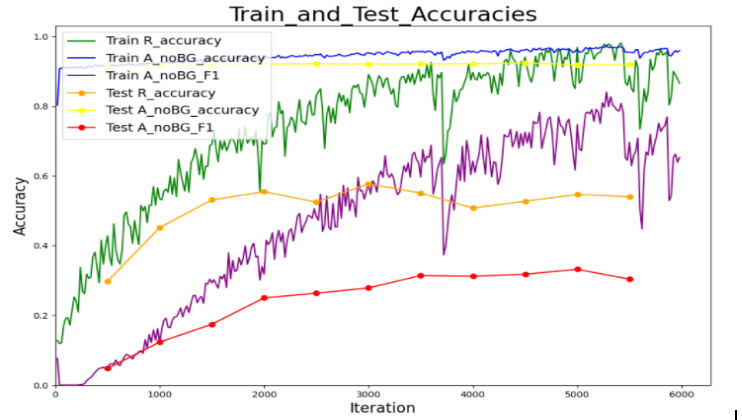}
\begin{figure}[h!]
\caption{Visualizations: accuracy on regularized model (best weight decay and best dropout values) on new set of features.}
\end{figure}

\includegraphics[scale=0.45]{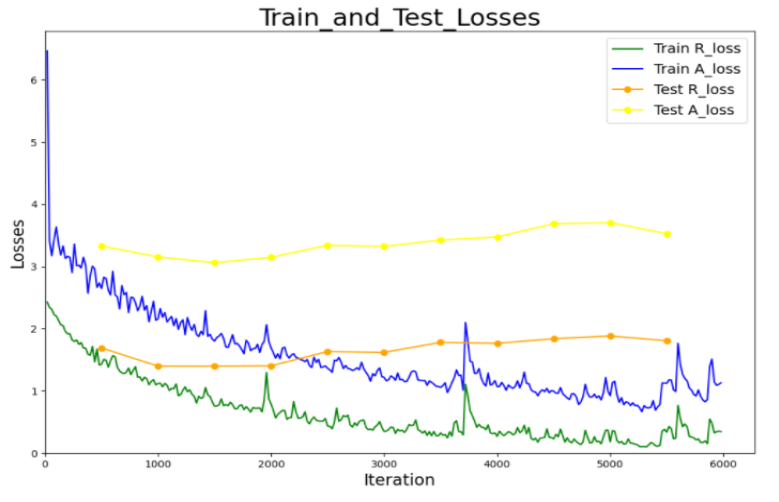}
\begin{figure}[h!]
\caption{Visualizations: loss on regularized model (best weight decay and best dropout values) on new set of features.}
\end{figure}

\includegraphics[scale=0.58]{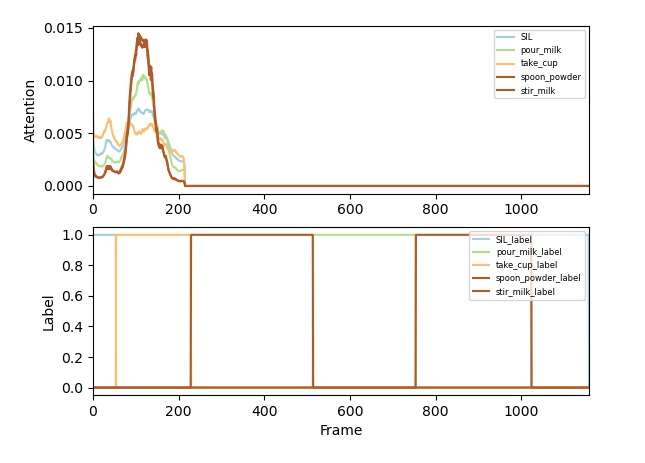}
\begin{figure}[h!]
\caption{Visualizations: attention mechanism on regularized model (best weight decay and best dropout values) on new set of features.}
\end{figure}

Gathering the above experimental results, we observe the following:
\begin{enumerate}
\item Initially when we had the subplots from \textit{Figure 9} on one plot, the magnitude of our attention was surprisingly small and illegible
\item After separating into two subplots (as shown in \textit{Figure 9)} our attention (subplot) mechanism seemed to concentrate only on the first part of a given test video, thus missing a lot of information on the rest of the video
\item Furthermore, different actions in a given test video seemed to correlate with each other as shown in the subplot attention \textit{Figure 9}, where the attention is both high and low within the same frame. This means that our attention focuses on the same location in a test video
\end{enumerate}

\section{Conclusion and Future Directions}
Based on our results and analysis, we propose the following next steps to help improve our model:

\renewcommand\labelitemii{$\blacksquare$}
\begin{itemize}
\item Change our network structure and/or add a penalization loss term
\item Penalize the attention such that it does not focus on the same video frame
\end{itemize}

Video learning under label constraints is a challenging task but with the huge traction it has gained in computer vision tasks, very promising indeed.\\

\bibliographystyle{unsrt}  


\end{document}